\documentclass{article}

\usepackage[final,nonatbib]{neurips_ts4h_2022}

\usepackage[utf8]{inputenc} 
\usepackage[T1]{fontenc}    
\usepackage{hyperref}       
\usepackage{url}            
\usepackage{booktabs}       
\usepackage{amsfonts}       
\usepackage{nicefrac}       
\usepackage[dvipsnames]{xcolor}         

\usepackage{graphicx}
\usepackage{subcaption}
\usepackage{wrapfig}
\hypersetup{
    colorlinks=true,
    linkcolor=RoyalBlue,
    filecolor=magenta,
    urlcolor=RoyalBlue,
    citecolor=RoyalBlue
}

\title{Temporal patterns in insulin needs for Type 1 diabetes}

\author{
  Isabella Degen\\
  Department of Computer Science\\
  University of Bristol\\
  Bristol, BS8 1TH\\
  \texttt{isabella.degen@bristol.ac.uk} \\
  \And
  Zahraa S. Abdallah \\
  Department of Engineering Mathematics \\
  University of Bristol\\
  Bristol, BS8 1TH\\
  \texttt{zahraa.abdallah@bristol.ac.uk} \\
}

\begin{document}

\maketitle

\begin{abstract}
Type 1 Diabetes (T1D) is a chronic condition where the body produces little or no insulin, a hormone required for the cells to use blood glucose (BG) for energy and to regulate BG levels in the body. Finding the right insulin dose and time remains a complex, challenging and as yet unsolved control task. In this study, we use the OpenAPS Data Commons dataset, which is an extensive dataset collected in real-life conditions, to discover temporal patterns in insulin need driven by well-known factors such as carbohydrates as well as potentially novel factors. We utilised various time series techniques to spot such patterns using matrix profile and multi-variate clustering. The better we understand T1D and the factors impacting insulin needs, the more we can contribute to building data-driven technology for T1D treatments.
\end{abstract}

\section{Introduction} \label{Introduction}
 Patients with T1D need to manually inject insulin to keep alive. Despite advances in treatments for T1D, it is challenging to maintain glucose levels which keep patients safe from complications. A considerable part of the research focuses on automated insulin delivery systems (AID), also called artificial pancreas (APS) or closed loop systems (CL) \cite{Contreras2018}. These systems use an insulin pump to deliver the insulin and a continuous glucose monitoring sensor (CGM) as a signal to be regulated \cite{Bertachi2018}. Research into these systems focuses on the safety of the system, improving the algorithms, accurately predicting blood glucose and predicting system failures such as insulin infusion set and blood glucose sensor failures. 

This study, on the other hand, is using data produced by an AID to find temporal patterns in insulin needs that are not yet known. The data studied stems from the open-source automated insulin delivery (OSAID)\cite{OpenAPS} systems, see appendix \ref{sec:dataset-access}. OSAIDs automatically adjust insulin based on blood glucose levels from a CGM sensor. The automated nature of the data collection creates a dataset that is more complete and accurate than other datasets where people have to manually record their treatment (which is prone to error and missing values). However, this data collected in real-life conditions comes with its own challenges of irregularities, non-even sampling, and missing data, which makes it hard to be handled with various standard time series techniques. 

In this paper, we implement intensive pre-processing on the OpenAPS Data Commons dataset in order to utilise time series techniques to find temporal patterns in insulin need linking insulin need to insulin on board (IOB), carbohydrates on board (COB) and blood glucose (BG). The rest of this paper is organised as follows: section \ref{literature-review} gives a review of the literature, section \ref{method} describes the method applied in this paper, and section \ref{results} lists the results achieved by the different techniques. Finally, section \ref{conclusion} concludes the research and lists directions for future work.

\section{Related work}\label{related work}
\label{literature-review}
Machine learning (ML) approaches are adopted to address the complex and non-linear dynamics of regulating BG using insulin. \cite{Contreras2018} gave examples of problems addressed by machine learning such as blood glucose prediction, overcoming shortcomings of deterministic automated insulin delivery algorithms such as predicting set failure \cite{Cescon2016} or sensor failure \cite{Turksoy2017} and incident prediction such as predicting hypoglycemia \cite{San2016}. Another research area aims to create new medical knowledge which focuses on issues such as predicting diabetes onset \cite{Khan2021}, predicting changes in behaviour, predicting BG and evaluating the efficacy of treatment. The researchers stress the importance of data-driven methods and the shift towards tailored management of therapies. 

Existing research employs a wide range of ML methods, Support vector machines (SVM) are widely used algorithms \cite {Kavakiotis2017}. Random forest was used for blood glucose prediction using a multivariate dataset to capture daily rhythms in glucose metabolism by introducing a time feature into the prediction \cite{Georga2012}. Other researchers attempted to predict blood glucose using a combination of methods: genetic programming, random forests, k-nearest neighbours, and grammatical evolution \cite{Hidalgo2017}. Data used for the aforementioned studies are collected in clinical settings with a small sample of patients (around 5-30 people). Also, the focus is on prediction rather than finding temporal patterns. 

Open-source automated insulin delivery has been extensively and independently studied \cite{Weisman2017,Donnell2019,Knoll2022,Braune2019,Petruzelkova2018,LEWIS2018,Oliver2019}.
While the OpenAPS Data Commons dataset has been studied to investigate clinical outcomes and the safety of the system it has not yet been used as a source for learning more about T1D, or changing current treatment advice. One study, while still mainly looking at the blood glucose data, has gone deeper into this dataset looking at blood glucose variability outcomes with regard to demographics \cite{Shahid2022}. This paper, however, looks at temporal patterns in insulin need that later on could be further studied for correlation and causality. To our knowledge, this is the first attempt to study this dataset with the purpose of generating new knowledge about T1D by identifying temporal patterns of insulin need using time series methods beyond the well-known pattern that eating more carbohydrates needs more insulin.

\section{Method} \label{method}
The focus of this paper is to identify temporal patterns that would suggest something else than carbohydrates was driving the need for insulin. We utilised various time series techniques such as heatmaps, Matrix Profile (MP) \cite{Michael2016} and time series clustering. The patterns are identified by visual inspection of the results of these techniques. The patterns we looked for were times when insulin on board (IOB) did not rise or drop with carbohydrates on board (COB), times when blood glucose (BG) was rising without COB rising and times when BG was dropping without IOB rising.

The data used from the OpenAPS Data Commons datasets were the system logs that record the changes in the system. We focused on IOB, COB and BG. The "onboard" values IOB and COB are calculations that the OSAID performs to establish how much insulin respectively carbohydrates are acting at any point in time after insulin was injected respectively carbohydrates were eaten.

Each of the techniques used in this paper required some preprocessing of the data namely: making the time stamps uniform, dealing with missing data, making the time series regular (for MP and clustering) and z-score normalising the data for MP respectively min-max scaling the data for clustering.
The time stamps were made uniform by translating the various different formats to UTC. Time stamps without time zone information were imputed with time zone information from previous time stamps and missing time stamp entries were dropped. The irregular time series were resampled into regular time series with a frequency of a reading per hour respectively a reading per day aggregating multiple values using mean, min, max and std. To avoid resampling over periods without sufficient data points a maximum allowed gap between two readings was enforced. For the hourly frequency, we required a minimum of one reading for each hour of the day; for the daily frequency, we required a minimum of one reading every three hours of the day. Time periods with less frequent readings were excluded. See appendix \ref{sec:method-datapreprocessing} for more details on data preprocessing.
We investigated different time periods that are of interest from a domain perspective: months, weeks and days. For months and weeks, we used the resampled series of mean values for each day. For days we used the resampled series of mean values for each hour of the day. Each time period is expected to highlight different patterns of insulin needs. Looking at months could show patterns due to seasons (winter vs summer). Looking at weeks could show patterns of days with different activity levels (workday/weekend). Finally, looking at days could show different eating/sleeping/activity patterns. 

To validate that the data preprocessing did not significantly change the distribution of the data and or introduced patterns that weren't originally in the data, the statistical properties of each of the different time series were calculated and visualised in box and violin plots and compared with each other. Furthermore, we calculated the confidence intervals for the mean values of the regular, raw and regular, hourly and daily time series. The different time resolutions of the time series are expected to make certain patterns more obvious and hide others but they must not contradict each other. See Appendix \ref{sec:preprocessing-evaluation}.

First, heatmaps were used to visualise the time series data for three random patients across days of the week and months of the year.

Second, the Matrix Profile (MP) was used to find motifs and discords in the time series data across weeks. MP is an efficient algorithm for finding signatures in time series and requires a window size $m$. MP works by sliding a window of size $m$ across the whole time series (or across multiple time series for the multivariate version) and calculating the distance between this window and each of the subsequent overlaps excluding the window itself to avoid trivial matches. For each point in time, the minimal distance gets recorded resulting in a new time series of the minimal distance for each point in time of the time series. The resulting minimal distance time series can then be used to find motifs and discords in the original time series. A subsequence of length $m$ starting at the index of the lowest minimal distance is a motif, while a subsequence of length $m$ starting at the index of the biggest minimal distance is a discord in the original time series. Window size is usually obtained through domain knowledge and hugely impacts the motifs and discords found. We calculated the Matrix profile for the longest consecutive time series with the frequency of a mean value per day. The window size was set to $m=7$, to indicate a period of seven consecutive days. The reason for choosing a week as motif length was to attempt to find patterns in weeks such as weeks with lower or higher insulin on board caused by the female hormone cycle.

Third, k-means was used to group the time series into clusters of similar days/weeks of IOB, COB and BG patterns. The resulting clusters were analysed for how they differ from each other and if we can identify potentially unknown patterns in insulin need by visually inspecting the barycenter of these clusters. Barycenters are the global average of a collection of time series. They can be calculated using different methods. Given we used dynamic time warping (DTW) as distance measure\cite{Berndt1994} it is then common practice to use DBA \cite{Petitjean2011} as barycenters. We clustered the hourly time series by IOB, COB and BG alone but always visualised all three different time series for the same day/week. We also ran multivariate time-series k-means clustering of IOB, COB and BG.
We found the number of clusters $k$ using silhouette analysis and the elbow method. We also used silhouette analysis to assess the impact of the quality of the clusters by limiting dynamic time warping using a Sakoa Chiba\cite{Sakoe1978} band. Finally, to account for noise and bias in the data we cross-validated the patterns found by splitting the data into $n=11$ folds and calculating multivariate k-means clustering of IOB, COB and BG $n$ times leaving one of the folds out each time.

\section{Results} \label{results}
\begin{figure}[ht]
    \centering
    \includegraphics[width=0.85\textwidth]{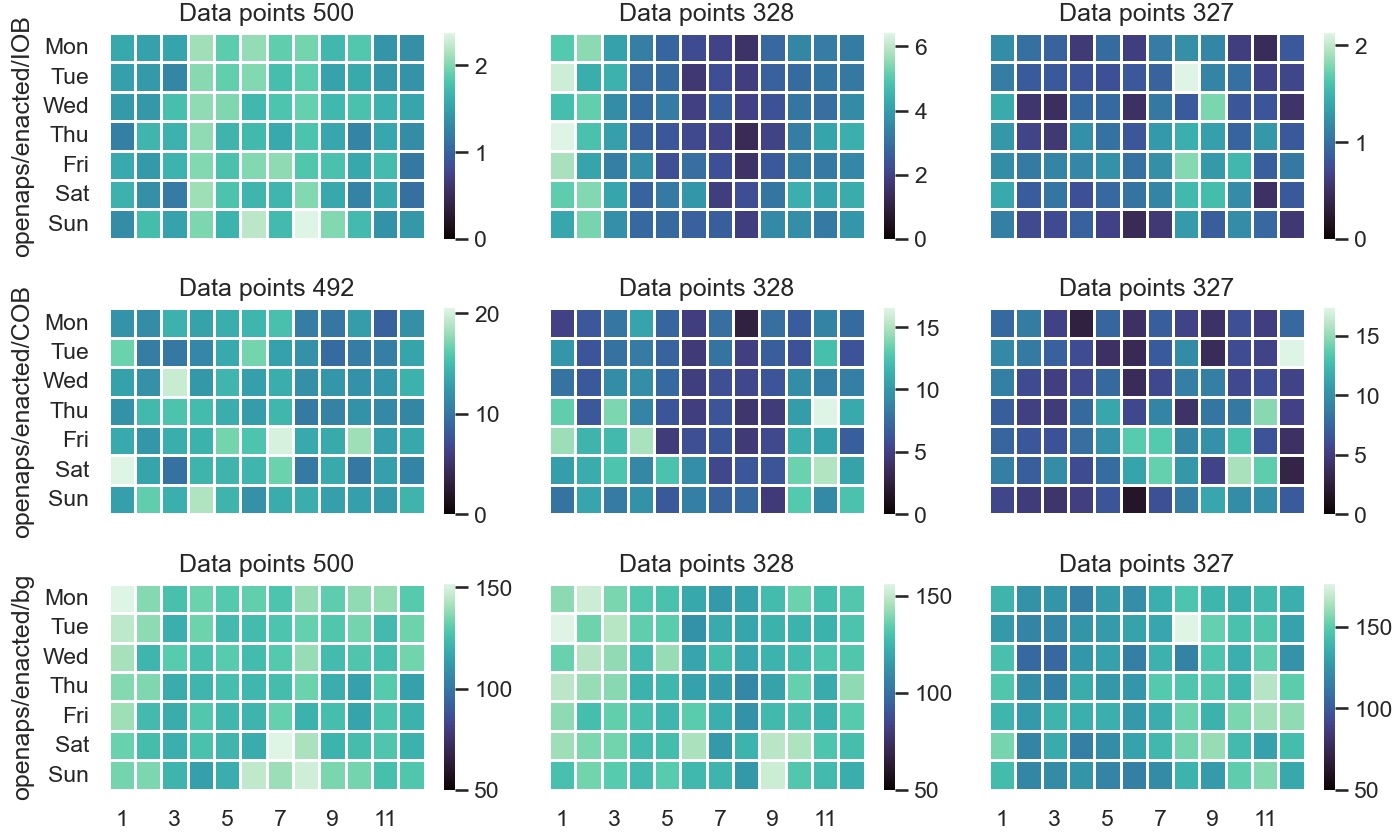}
    \caption{Heatmap for IOB, COB and BG as rows and P1, P2 and P3 as columns. The y-axis of each subplot plot is the days of the week and the x-axis is the months of the year using resampled time series with a mean value for each day.}
    \label{heathmaps-months}
\end{figure}

The heatmaps of IOB, COB and BG for months of the year (x-axis) and days of the week (y-axis) in figure \ref{heathmaps-months} show that for P1 (first column in the plot) the months 4-8 have a higher IOB without COB being higher. For P2 (second column) months 1 and 2 show a higher IOB without COB showing the same pattern. For P3 (third column) there's less of a pattern in IOB and COB but BG seems to be higher in months 1-6. These patterns suggest that some seasonal changes seem to be happening and that they vary from individual to individual.

\begin{figure}[h]
     \centering
     \begin{subfigure}[b]{0.49\textwidth}
         \centering
         \includegraphics[width=\textwidth]{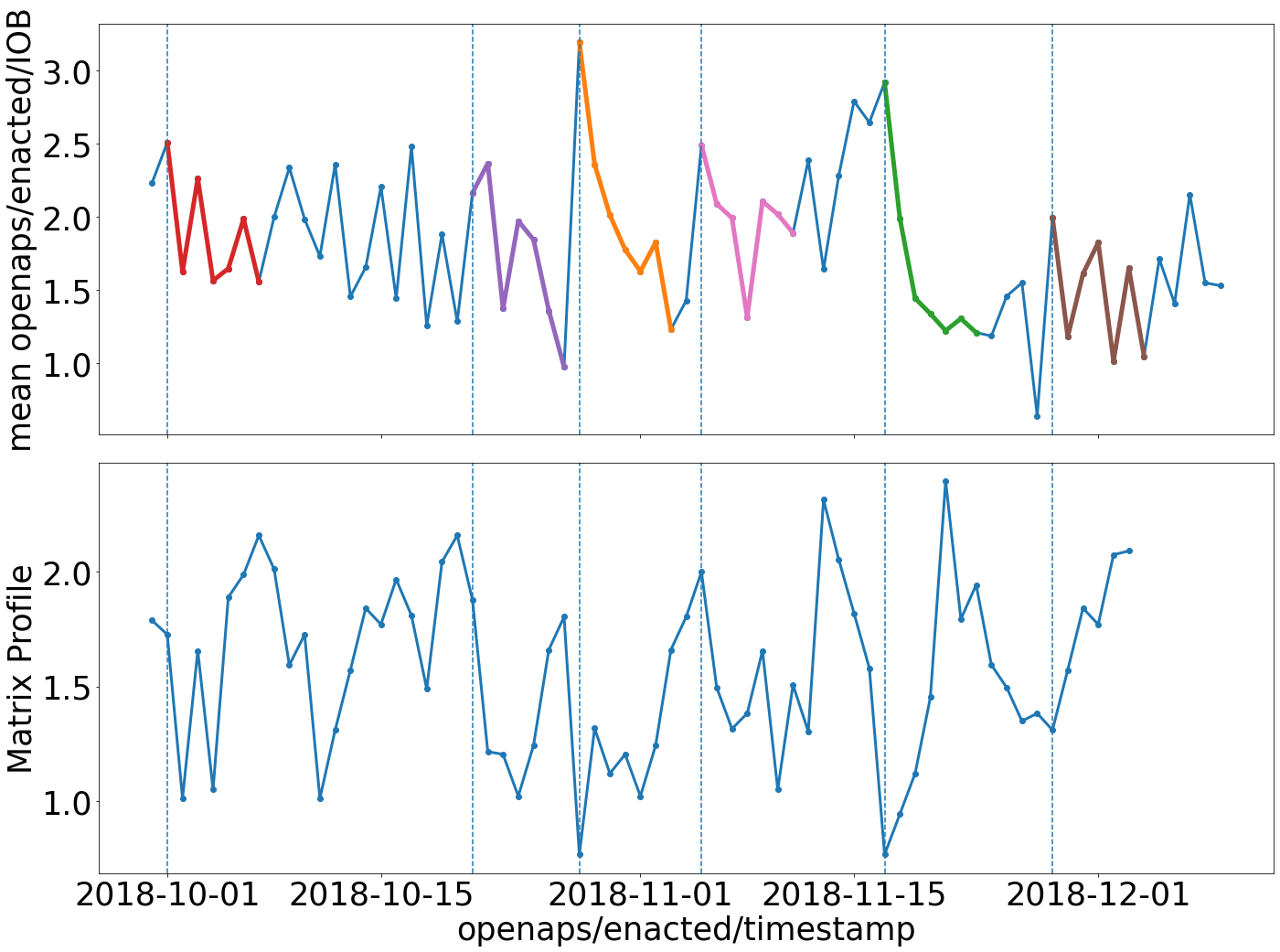}
         \caption{Top motifs, distance $\leq2.33$}
     \end{subfigure}
     \hfill
     \begin{subfigure}[b]{0.49\textwidth}
         \centering
         \includegraphics[width=\textwidth]{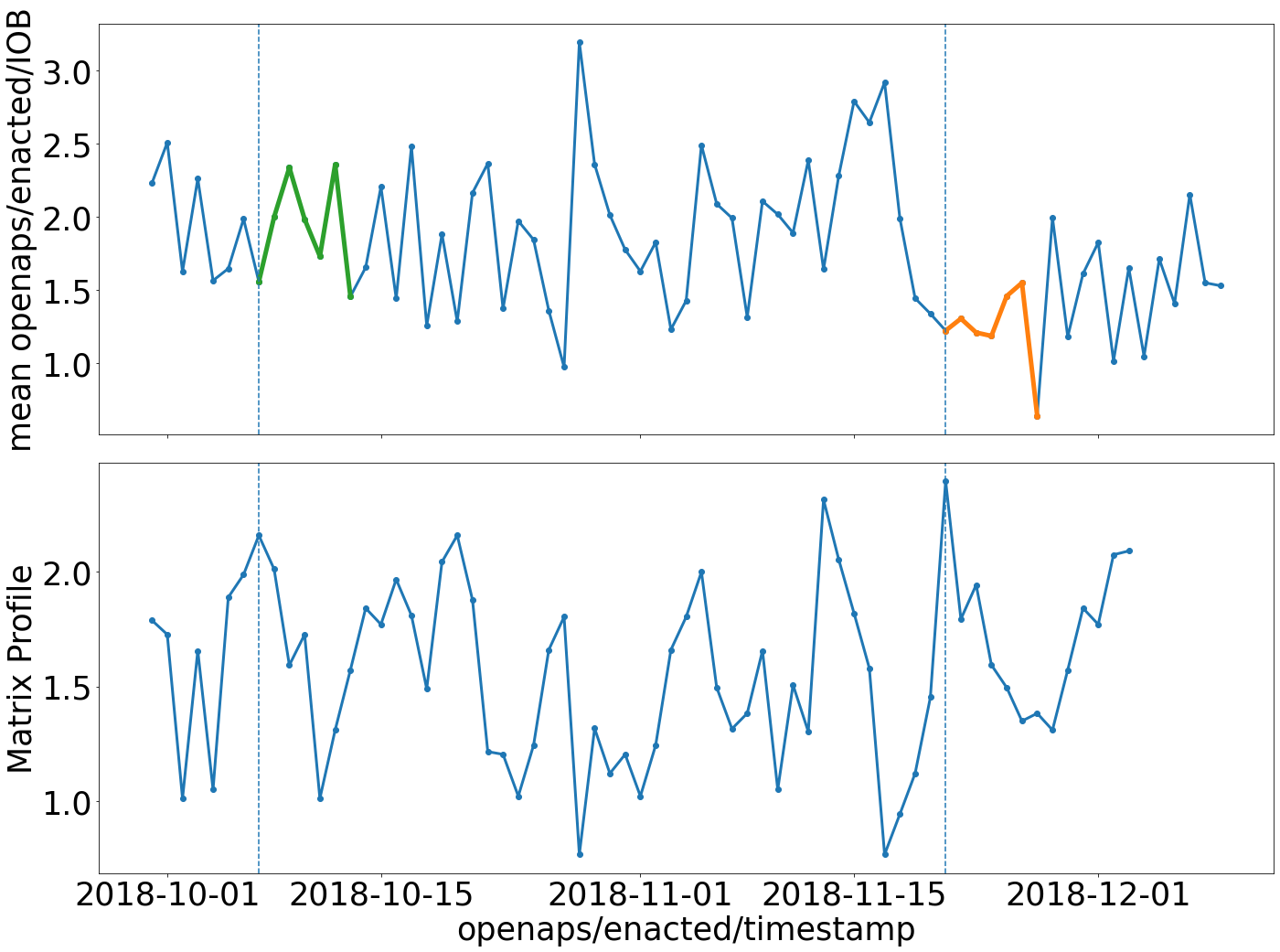}
         \caption{Top discord, distance $2.4$}
     \end{subfigure}
        \caption{(a) Top motifs and (b) top discord on top and matrix profile at the bottom for longest continuous regular time series of IOB for P1. Window size is $m=7$, the length of the time series is $71$ days and resampled as daily mean value.}
        \label{fig:mp-motifes}
\end{figure}

The matrix profile for the resampled IOB time series with a mean value for each day over the longest continuous time series is shown in figure \ref{fig:mp-motifes}. The window size was set to $m=7$, to indicate a period of seven consecutive days. Sub-figure (a) in \ref{fig:mp-motifes} shows the top motifs and (b) the top discord. Both show the matrix profile below the IOB time series. 
The top motifs are the 7 consecutive days that are most like each other. The top discords are 7 consecutive days which are the least like each other. The IOB discord shows that a pattern of a stable week followed by a drop towards the end of the seven days is a discord. While the matrix profile was not that insightful in finding patterns it did highlight that seven relatively constant days of IOB were a discord and that motifs had rather drastic ups and downs of IOB from day to day, suggesting that treatment needs to be adjusted dynamically from day to day.

The barycenters for k-means clustering of time series with hourly frequency clustered by (a) IOB, COB and BG, (b) IOB only and (c) BG only are shown in figure \ref{fig:barry-center-comparision-daily-ts}.
Unsurprisingly, all clusters show the three main meals as the most pronounced spikes. We can also recognise a long flat line, presumably the night, visible in all figures. Furthermore, each cluster shows the patterns we are looking for where IOB does not follow COB, BG rises without COB rising or drops without IOB rising. These are times when there was a higher/lower need for insulin than provided. Notably, all clusters show BG (green) rising for all three or some of the main meal spikes after COB (orange) has dropped. Furthermore, each of the sub-figures (a)-(c) has at least one cluster where BG rises in the evening/night ($\approx 4-7$ UTC) after IOB and COB from dinner have dropped. These rises happen alongside higher IOB, presumably provided by the OSAID to counteract the higher BG which is eventually successful. On top of the patterns visible in all clustering settings, there are also some patterns that are only visible for certain settings. The multivariate clustering (a) shows in cluster 2 a pattern where the BG rises in the evening/night followed by no/delayed breakfast($\approx 12$ UTC). Clustering by IOB only (b) shows in cluster 2 a pattern where BG rises in the evening/night followed by a breakfast with carbs for which no insulin was taken. It also shows that for lunch the insulin was then pre-injected. Finally, clustering by BG only (c) shows a pattern where BG rises before IOB and COB for a delayed breakfast ($\approx 15$ UTC, cluster 1 and cluster 2). Cluster 4 shows a pattern where IOB for what might be pre-lunch snacks ($\approx 16$ UTC) is much higher than COB.
The k-means clusters were not very tight with silhouette scores ranging from 0.081 (multivariate) to 0.189 (IOB). Using Sakoe Chiba bands reduced the quality of the clusters.

\definecolor{mblue}{HTML}{1f77b4}
\definecolor{morange}{HTML}{ff7f0e}
\definecolor{mgreen}{HTML}{2ca02c}
\begin{figure}[h]
    \centering
    \caption*{\fboxsep=2mm \fboxrule=1mm
            \fcolorbox{white}{mblue}{IOB }
            \fboxsep=2mm \fboxrule=1mm
            \fcolorbox{white}{morange}{COB} 
            \fboxsep=2mm \fboxrule=1mm
            \fcolorbox{white}{mgreen}{BG \hspace{3pt}}}
     \begin{subfigure}[b]{0.49\textwidth}
         \includegraphics[width=0.95\textwidth]{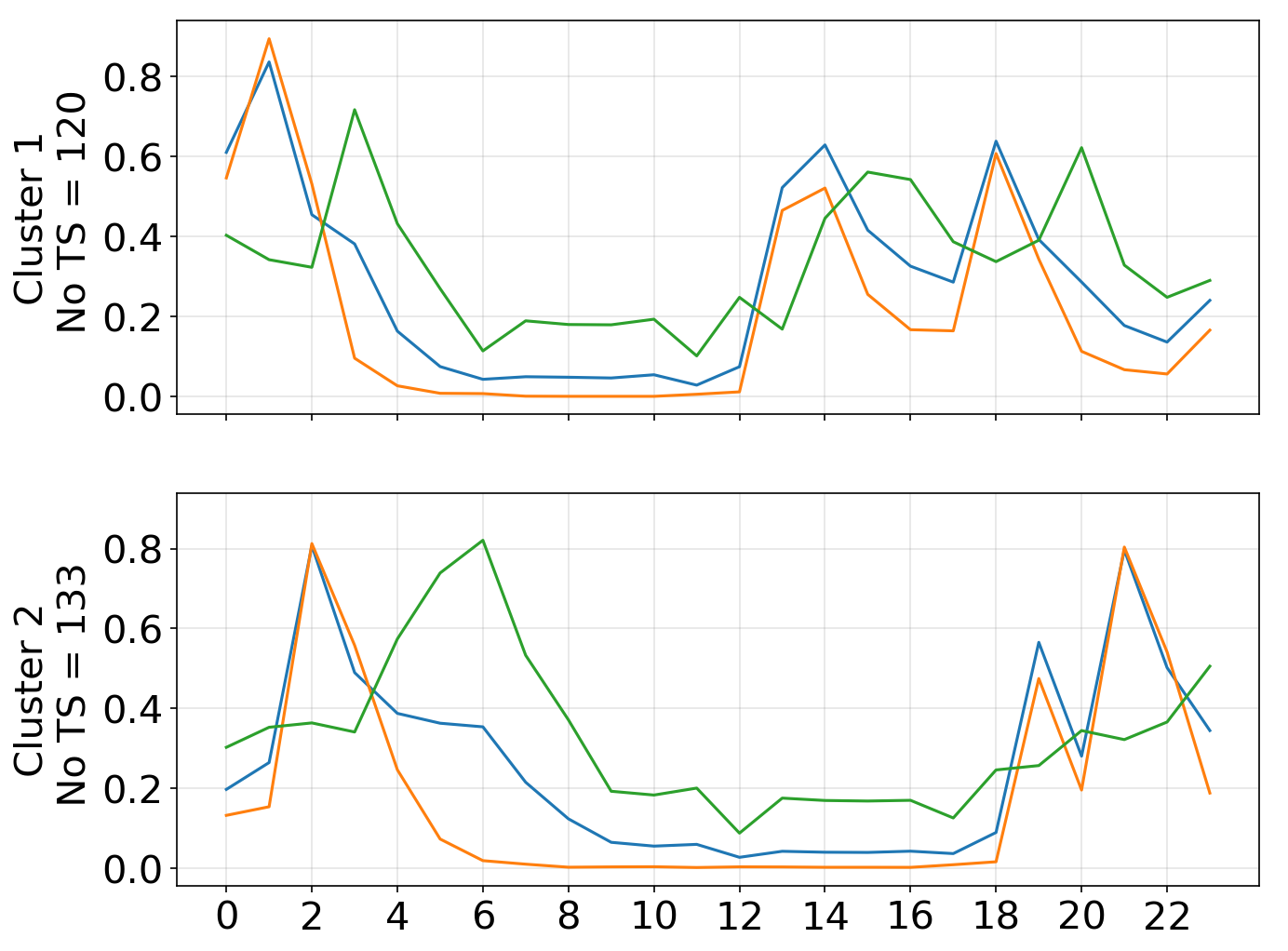}
        \caption{k-means clustered by IOB, COB and BG; $k=2$}
        \includegraphics[width=0.95\textwidth]{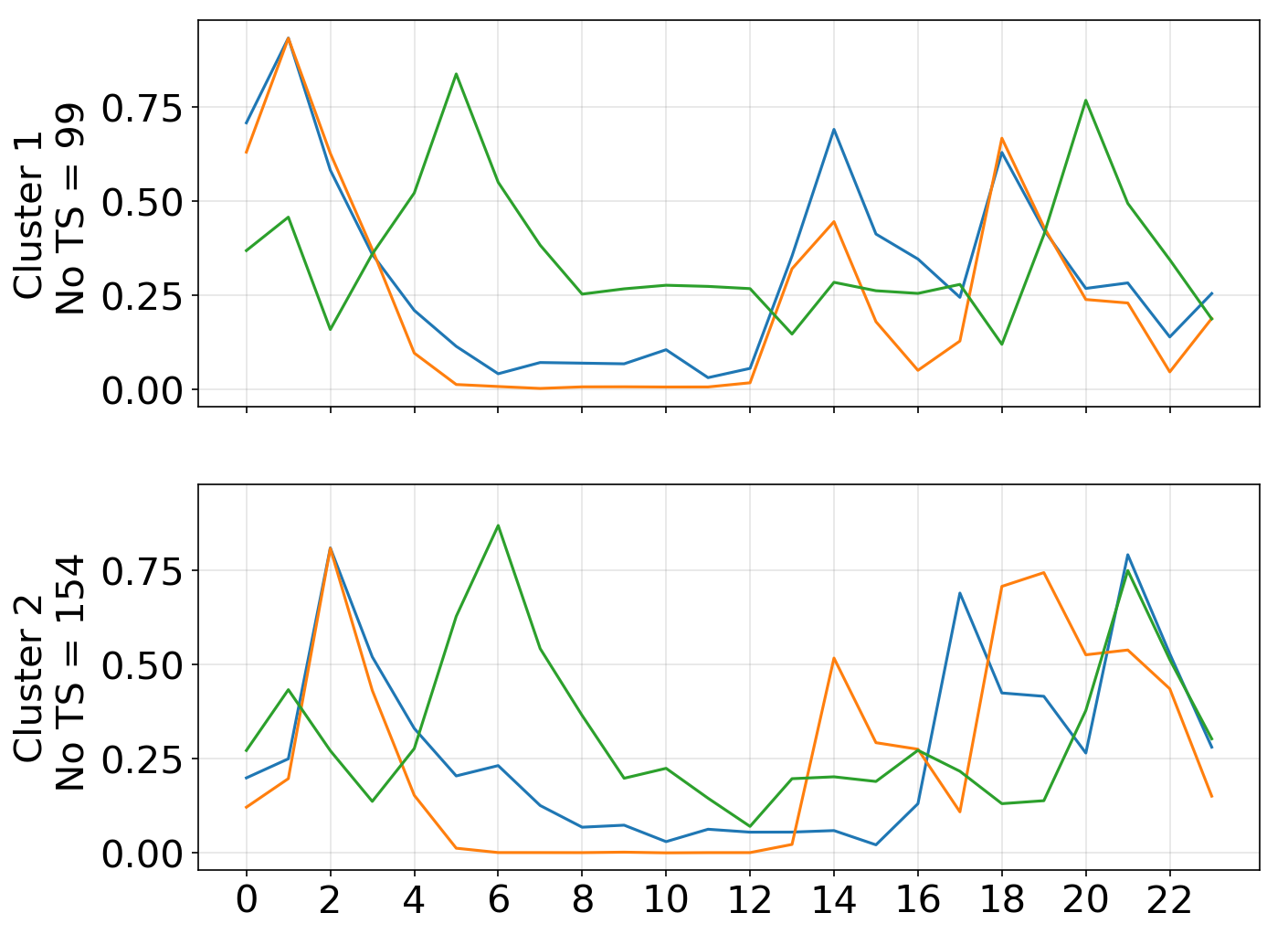}
        \caption{k-means clustered by IOB; $k=2$}
     \end{subfigure}
     \centering
     \begin{subfigure}[b]{0.49\textwidth}
         
         \centering
         \includegraphics[width=\textwidth]{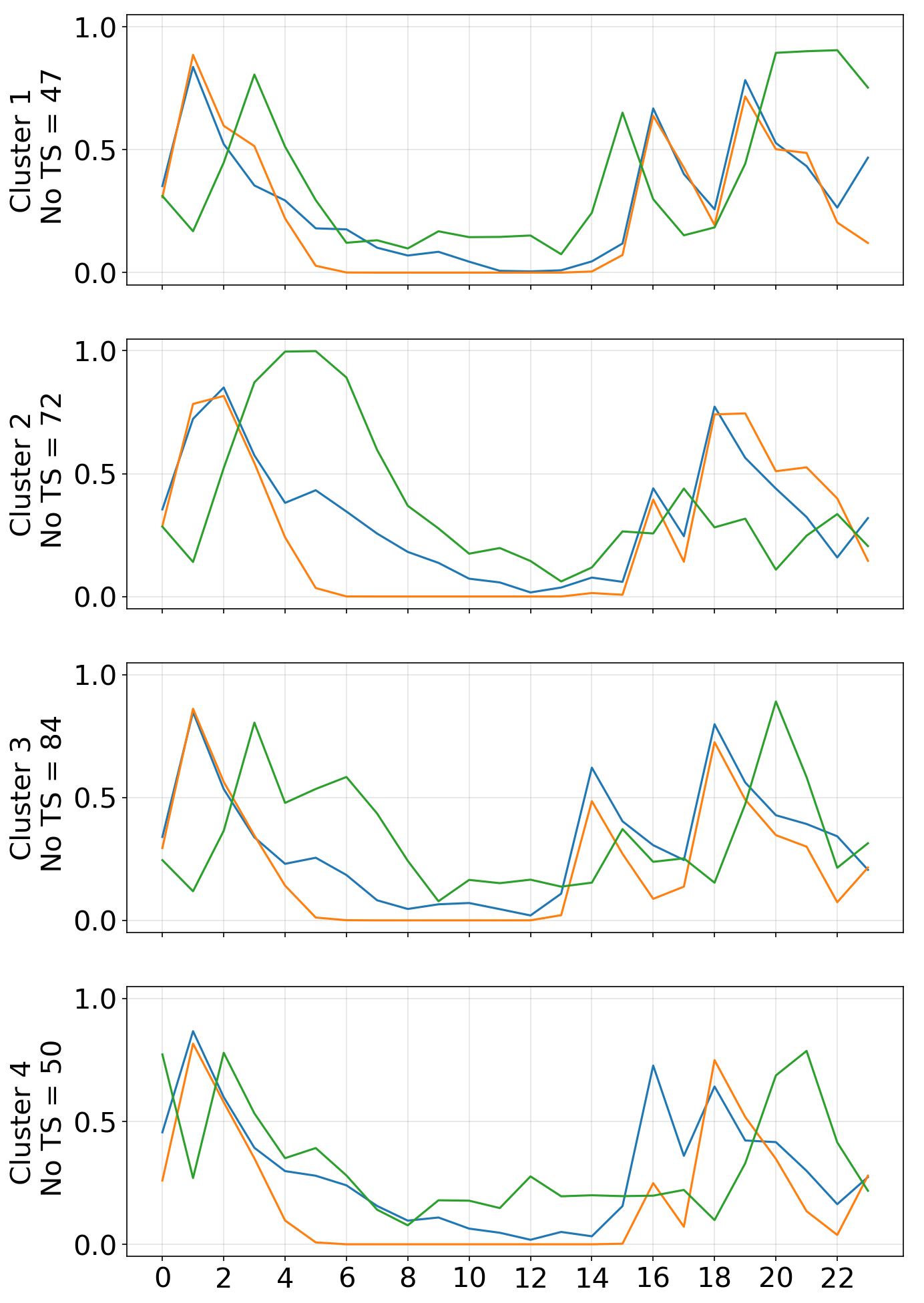}
         \par\bigskip
         \caption{k-means clustered by BG; $k=4$}
     \end{subfigure}
     
    \caption{Resulting barycenters for IOB, COB and BG when clustered (a) by all three, (b) IOB only (c) BG only. Clustered 253 days resampled into hourly mean values using time series k-means with DTW as a distance metric. For each sub-figure, the x-axis is the time of day in UTC and the y-axis is the min-max scaled mean values.}
    \label{fig:barry-center-comparision-daily-ts}
\end{figure}

The cross-validation for multivariate k-means clustering showed that all the patterns named above persisted for each of the 11 runs with each run dropping one of the folds of 23 days of data of the original 253 days of data for P1. Suggesting that the patterns we see are not simply due to noise or bias in the system, see \ref{sec:cross-validation}.

\section{Conclusion and future work \label{conclusion and future work}}
\label{conclusion}
We found in this paper interesting temporal patterns in insulin need that cannot be explained by carbohydrates themselves using existing time series techniques. If these patterns are driven by other known factors or go beyond what is currently known in medical science remains a question to be answered together with medical experts in T1D as well as the community of people with T1D using an OSAID system. While patterns found are very promising, they still lack scientific rigour and research into correlations and causalities that drive these patterns to truly be able to inspire new research into Type 1 Diabetes. 

\clearpage

\begin{ack}
We would like to thank UK Research and Innovation (UKRI) who is funding Isabella's PhD research through the UKRI Doctoral Training in Interactive Artificial Intelligence under grant EP/S022937/1. 

We are grateful to everybody involved in the Interactive AI CDT at Bristol University for their support and guidance.

We would like to thank Dana Lewis and the community who has tirelessly worked on the OSAID systems. We would also like to thank the OpenHumans platform and the people with Diabetes who have donated their data to research which formed the basis for this study.
\end{ack}

\bibliographystyle{unsrt}

\bibliography{neurips_ts4h_2022} 
\clearpage

\appendix
\section{Ethics statement}\label{sec:ethics-statement}
This study used the OpenAPS Data Commons dataset. The dataset consisted of system logs of the open-source automated insulin delivery system of n=183 people who have Type 1 Diabetes. The data has been voluntarily donated and made unidentifiable via the Open Humans platform. We have submitted an ethics application which has been reviewed and approved by the chair of the Faculty of Engineering Research Ethics Committee of the University of Bristol. The ethics approval code is 11270. The OpenAPS Data Commons research guidelines for working with the OpenAPS Data Commons dataset as set out by the community can be found here \cite{OpenAPSResearchGuidelines}.

\section{Access to dataset and code}\label{sec:dataset-access}
The OpenAPS Data Commons dataset was downloaded in April 2022 and consisted of data from n=183 individuals. The data stems from three different OSAID systems: OpenAPS, Loop and AndroidAPS. Access to the dataset needs to be requested. Instructions on how to apply can be found on the OpenAPS Data Commons website \cite{OpenAPSDataCommons}.

The code used to produce this paper can be found here \href{https://github.com/isabelladegen/insulin-need}{https://github.com/isabelladegen/insulin-need}.
The data cannot be shared publicly, therefore the notebooks showing the results cannot be shared either.
Example notebooks that show how to use the code will be made available.

\section{Data preprocessing}\label{sec:method-datapreprocessing}
\subsection{Uniforming time stamps}\label{sec:uniforming_time_stamps}
Due to many individuals having data in multiple time zones, it was decided to translate times to UTC to avoid jumps in time in the resulting time series data. This also provides better anonymisation and removes the perhaps tempting but incorrect assumption that people eat at a similar time of day and therefore time of day can be compared across individuals. 

Interpolation of missing time stamps from previous timestamps didn't seem sensible given this is system-written data, not user-entered data and therefore a gap in timestamps most likely means there was a problem with the system. Therefore we decided to exclude such times as for this paper we are not interested in patterns happening while the system is not fully working.

\subsection{Resampling time series}\label{sec:resampling}
In normal operation the OSAID systems receive an updated BG reading from the continuous blood glucose sensor every five minutes therefore resampling into hourly/daily mean, min, max, and std values was deemed reasonable. However, different issues such as communication interruptions, insulin refills, sensor changes and sensor warm-up periods and sensor errors, etc. lead to frequent gaps in the data. To avoid resampling over periods of time where such issues occurred a minimal number of readings in the original time series for that period of time was enforced as this paper was not interested in patterns happening during these times.

\subsection{Assessing impact of preprocessing on data distribution}\label{sec:preprocessing-evaluation}
While variations between the original, irregular and resampled time series with hourly or daily frequencies were expected, changes in patterns were not acceptable. For example, if IOB in January was higher than in February this had to be the case across all differently sampled time series. The mean, standard deviation, variance and numbers of outliers can vary but not the pattern itself.

\begin{figure}[ht]
     \centering
     \begin{subfigure}[b]{0.92\textwidth}
         \centering
         \includegraphics[width=\textwidth]{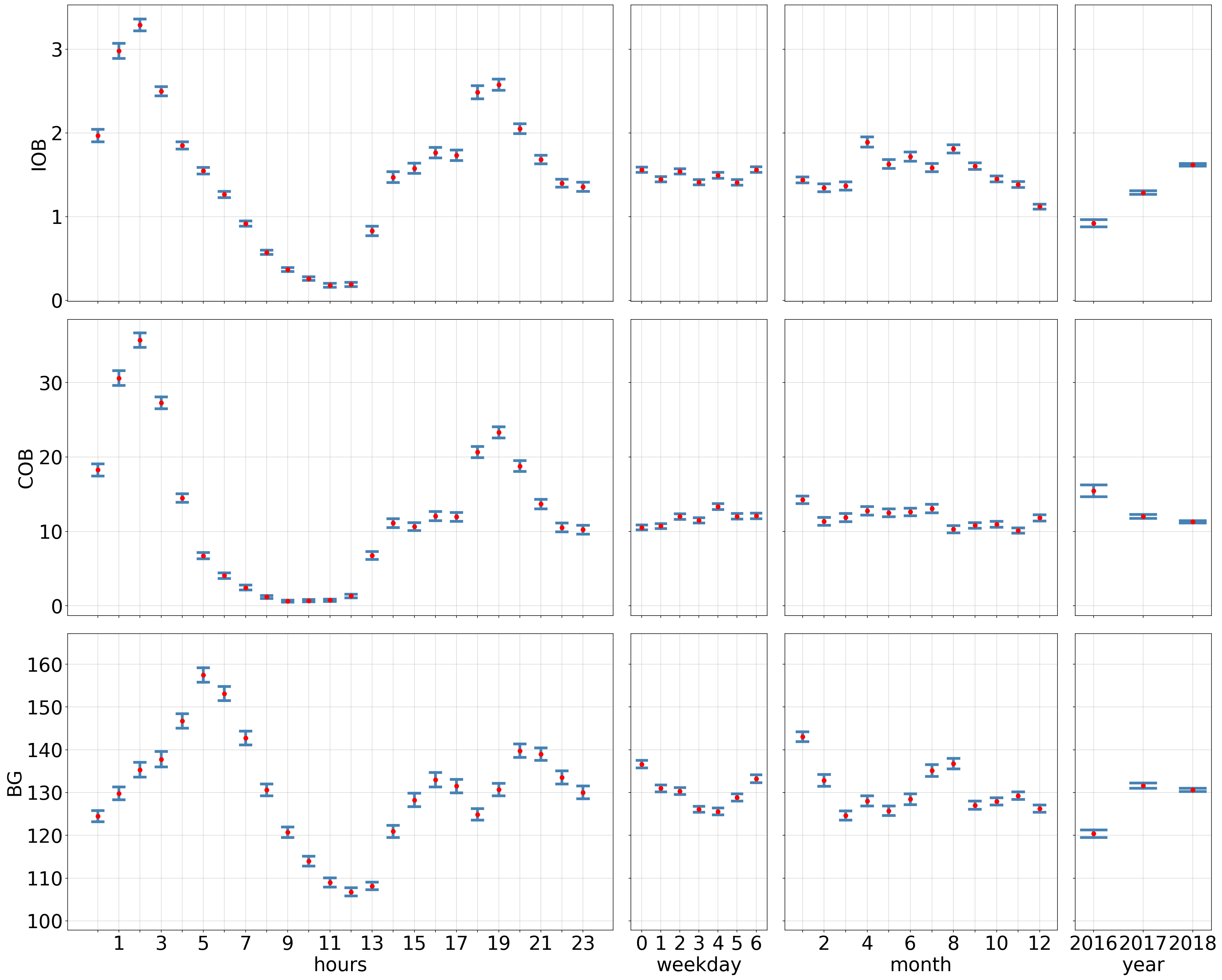}
         \caption{Original, irregular time series}
     \end{subfigure}
     \hfill
     \begin{subfigure}[b]{0.92\textwidth}
         \centering
         \includegraphics[width=\textwidth]{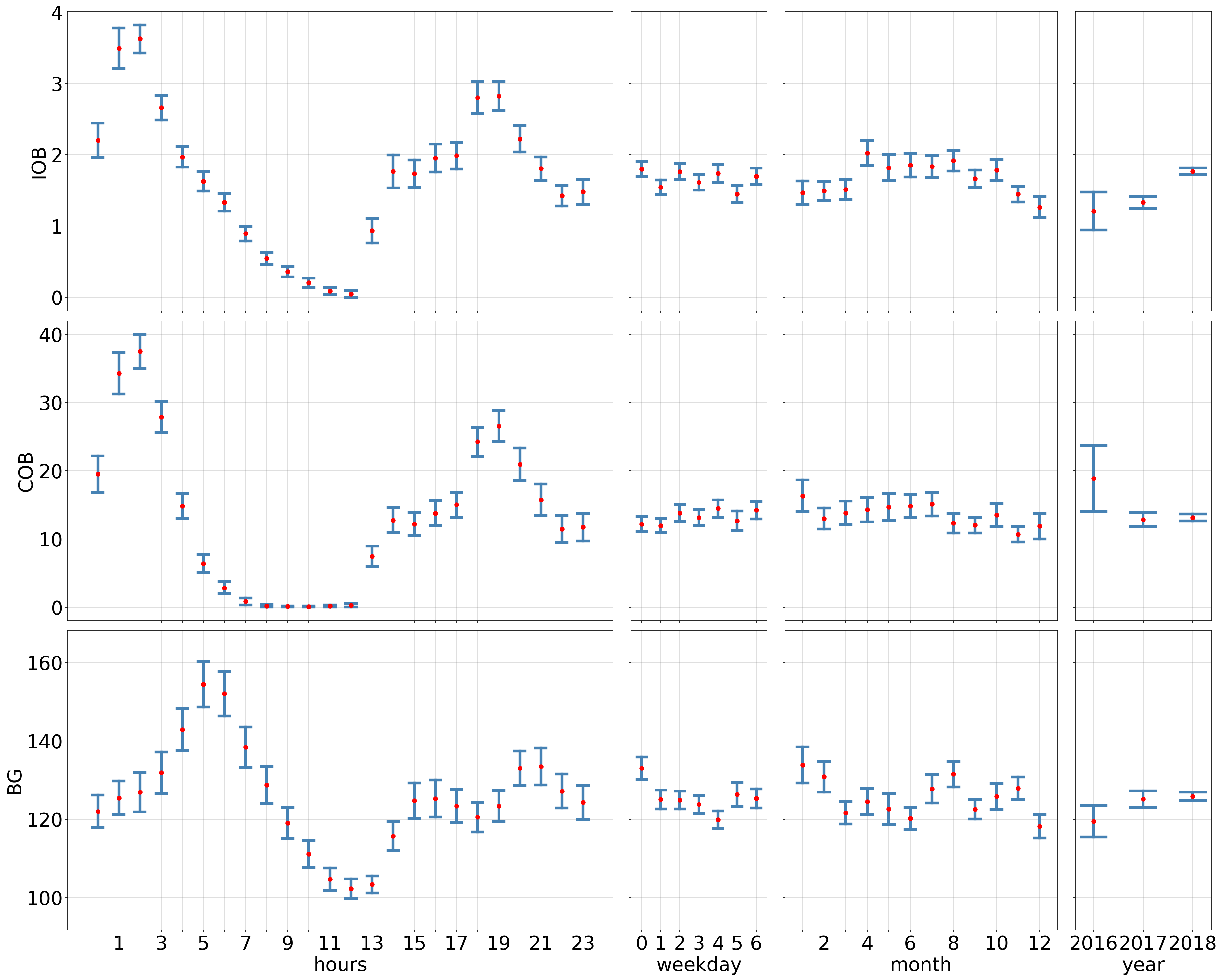}
         \caption{Resampled, regular time series with hourly frequency}
     \end{subfigure}
        \caption{Confidence intervals of mean values of IOB, COB and BG for P1 for hours, weekdays, months and years for (a) the original, irregular time series data and (b) the resampled, regular time series data where the resampling frequency was a reading per hour.}
        \label{fig:hourly-frequency-ci}
\end{figure}

\begin{figure}[h]
     \centering
     \begin{subfigure}[b]{0.92\textwidth}
         \centering
         \includegraphics[width=\textwidth]{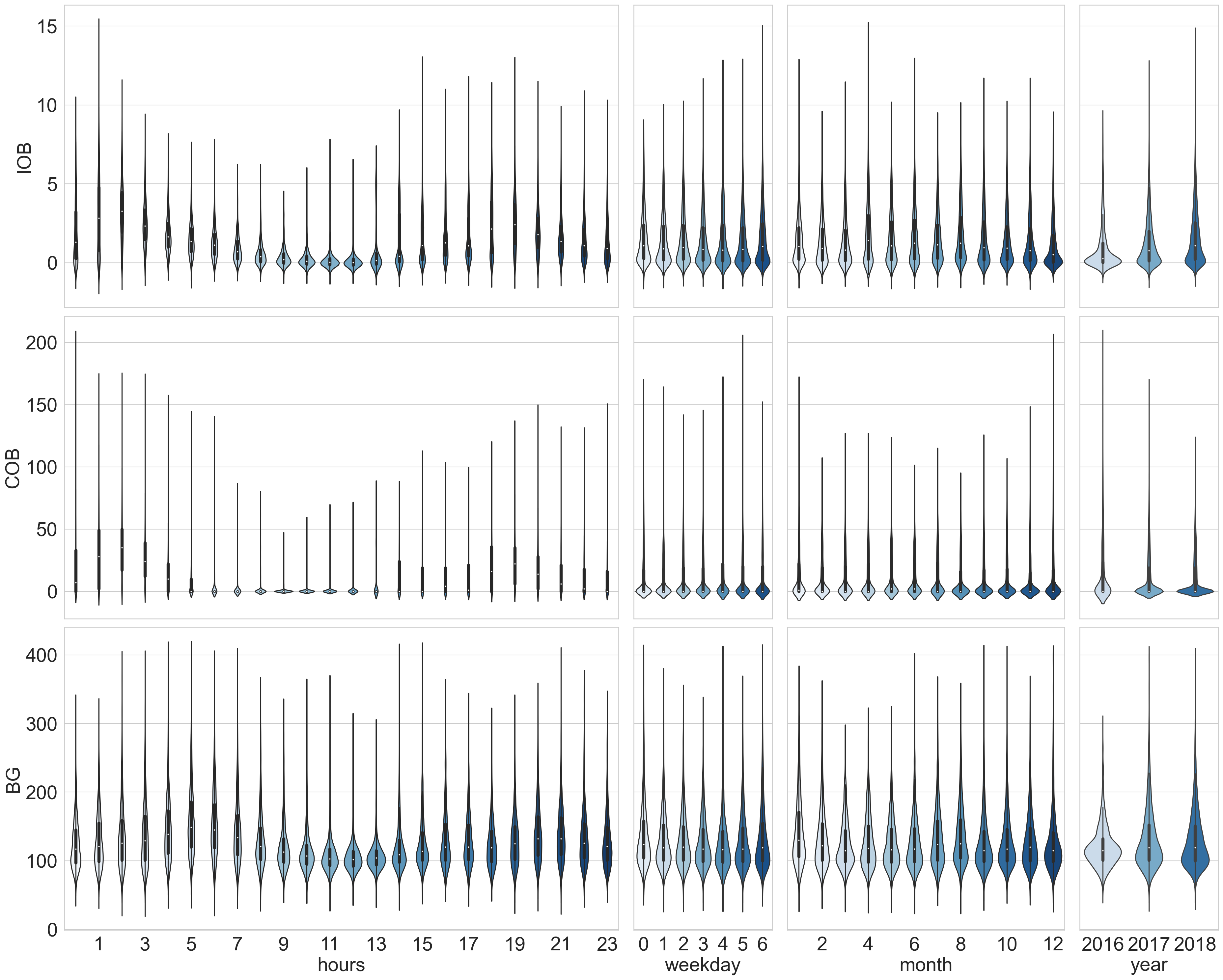}
         \caption{Original, irregular time series}
     \end{subfigure}
     \hfill
     \begin{subfigure}[b]{0.92\textwidth}
         \centering
         \includegraphics[width=\textwidth]{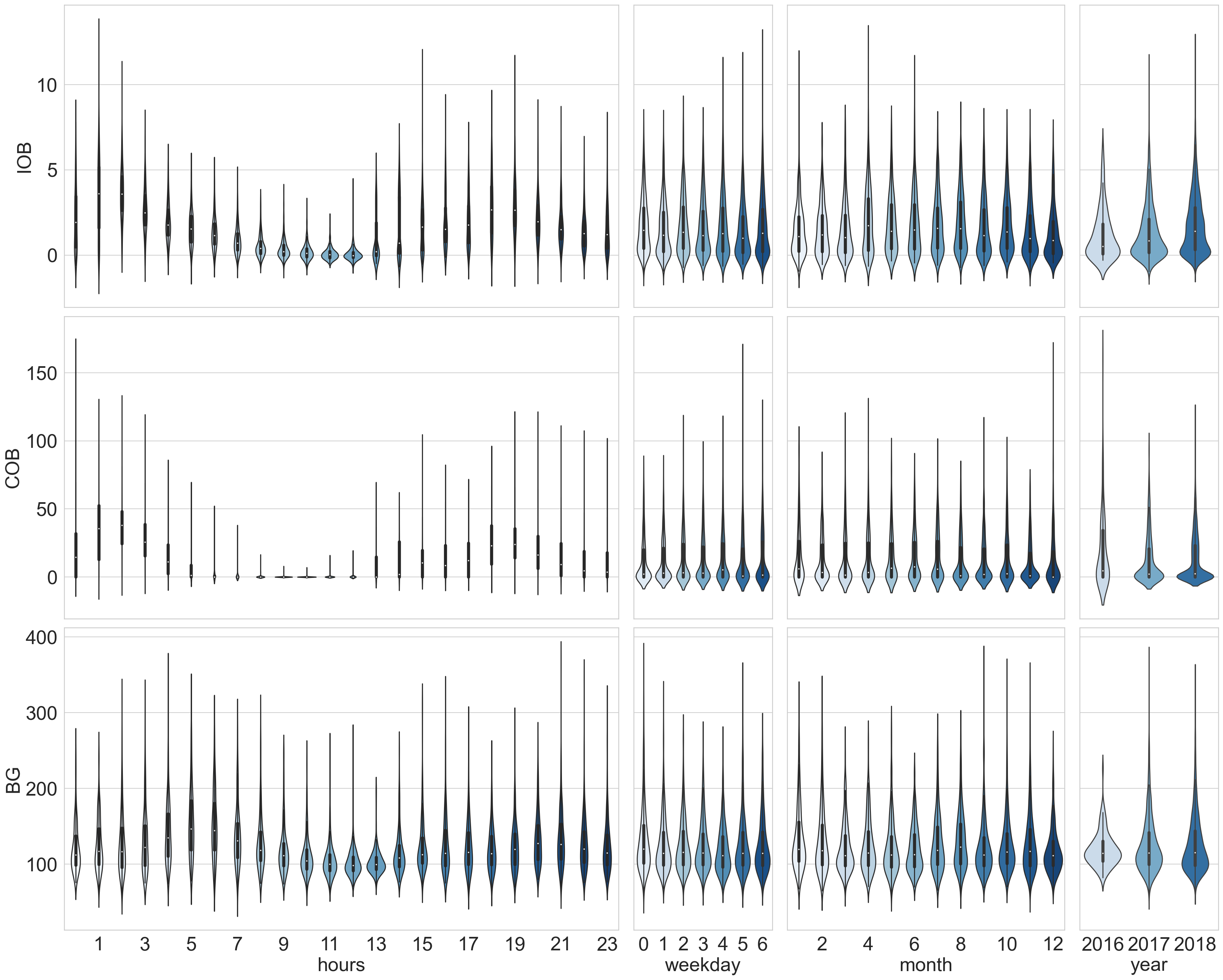}
         \caption{Resampled, regular time series with hourly frequency}
     \end{subfigure}
        \caption{Violin plots of mean values of IOB, COB and BG for P1 for hours, weekdays, months and years for (a) the original, irregular time series data and (b) the resampled, regular time series data where the resampling frequency was a reading per hour.}
        \label{fig:hourly-frequency-vp}
\end{figure}

\clearpage

\section{K-means cross-validation results}\label{sec:cross-validation}
\begin{figure}[h!]
    \centering
    \begin{subfigure}[b]{0.3\textwidth}
         \centering
         \includegraphics[width=\textwidth]{Figures/kmeans-multivariate-k2.png}
         \caption{All 253 days}
     \end{subfigure}
    \begin{subfigure}[b]{0.3\textwidth}
         \centering
         \includegraphics[width=\textwidth]{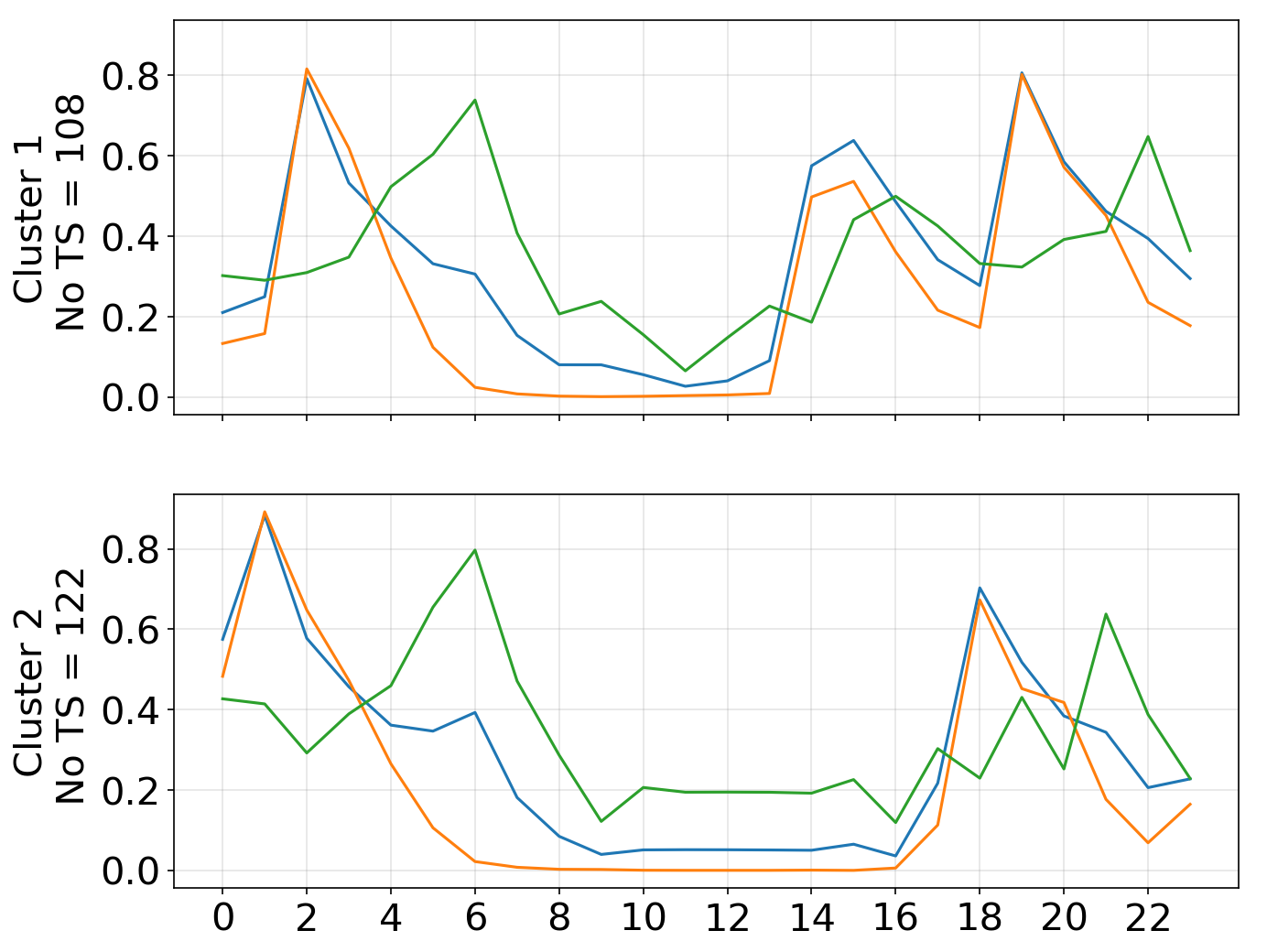}
         \caption{Fold 1}
     \end{subfigure}
     \begin{subfigure}[b]{0.3\textwidth}
         \centering
         \includegraphics[width=\textwidth]{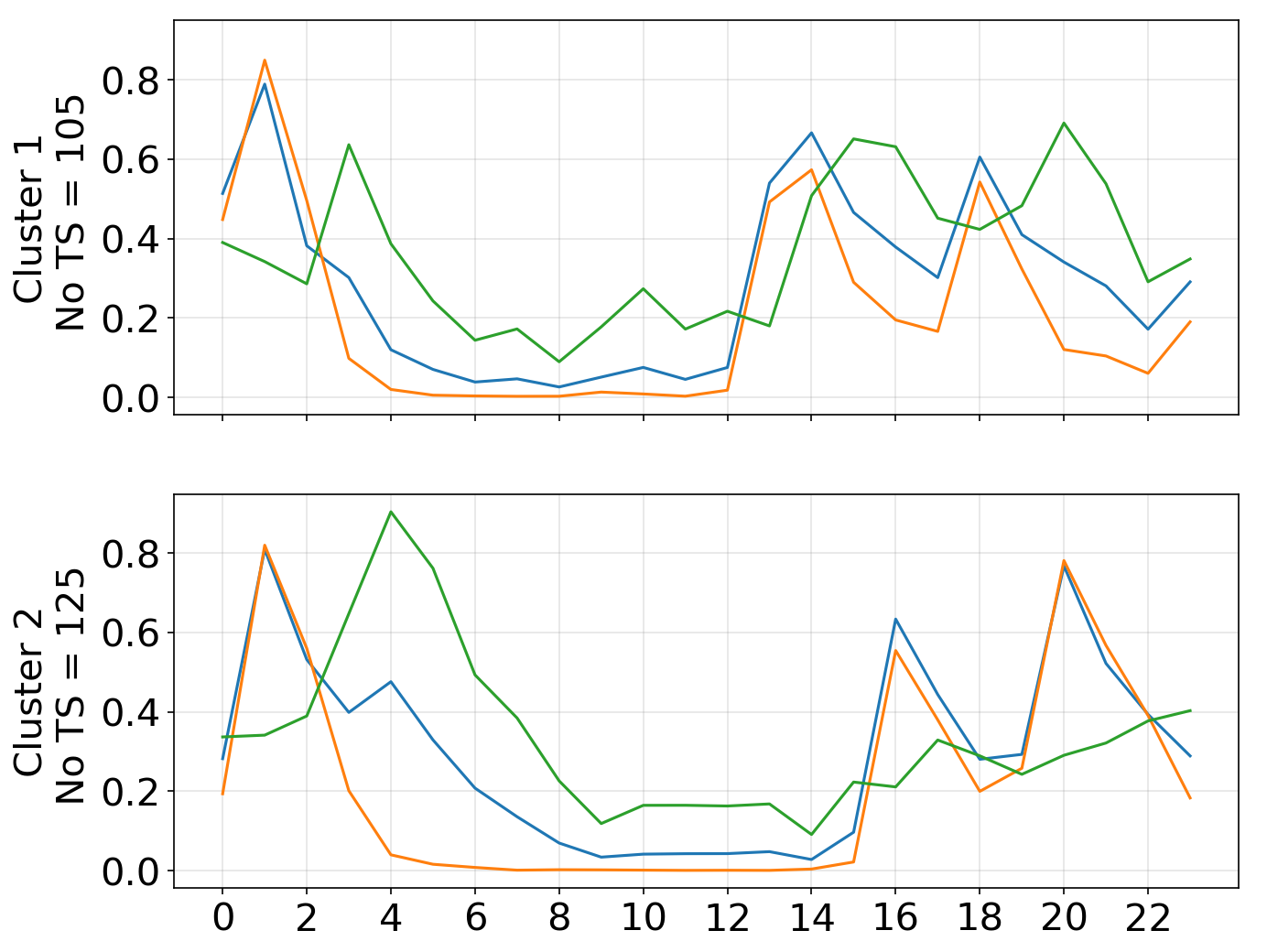}
         \caption{Fold 2}
     \end{subfigure}
     \begin{subfigure}[b]{0.3\textwidth}
         \centering
         \includegraphics[width=\textwidth]{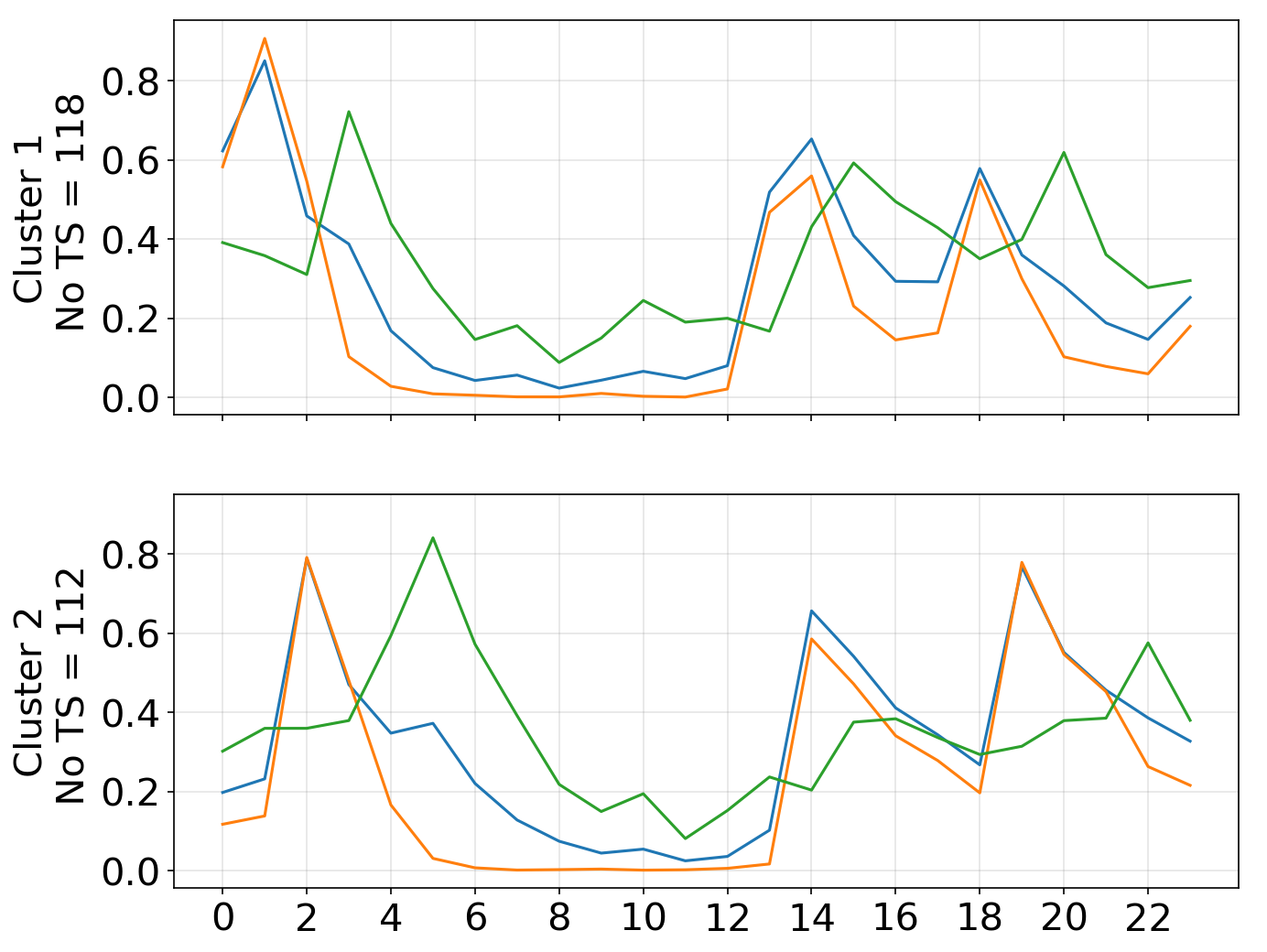}
         \caption{Fold 3}
     \end{subfigure}
     \begin{subfigure}[b]{0.3\textwidth}
         \centering
         \includegraphics[width=\textwidth]{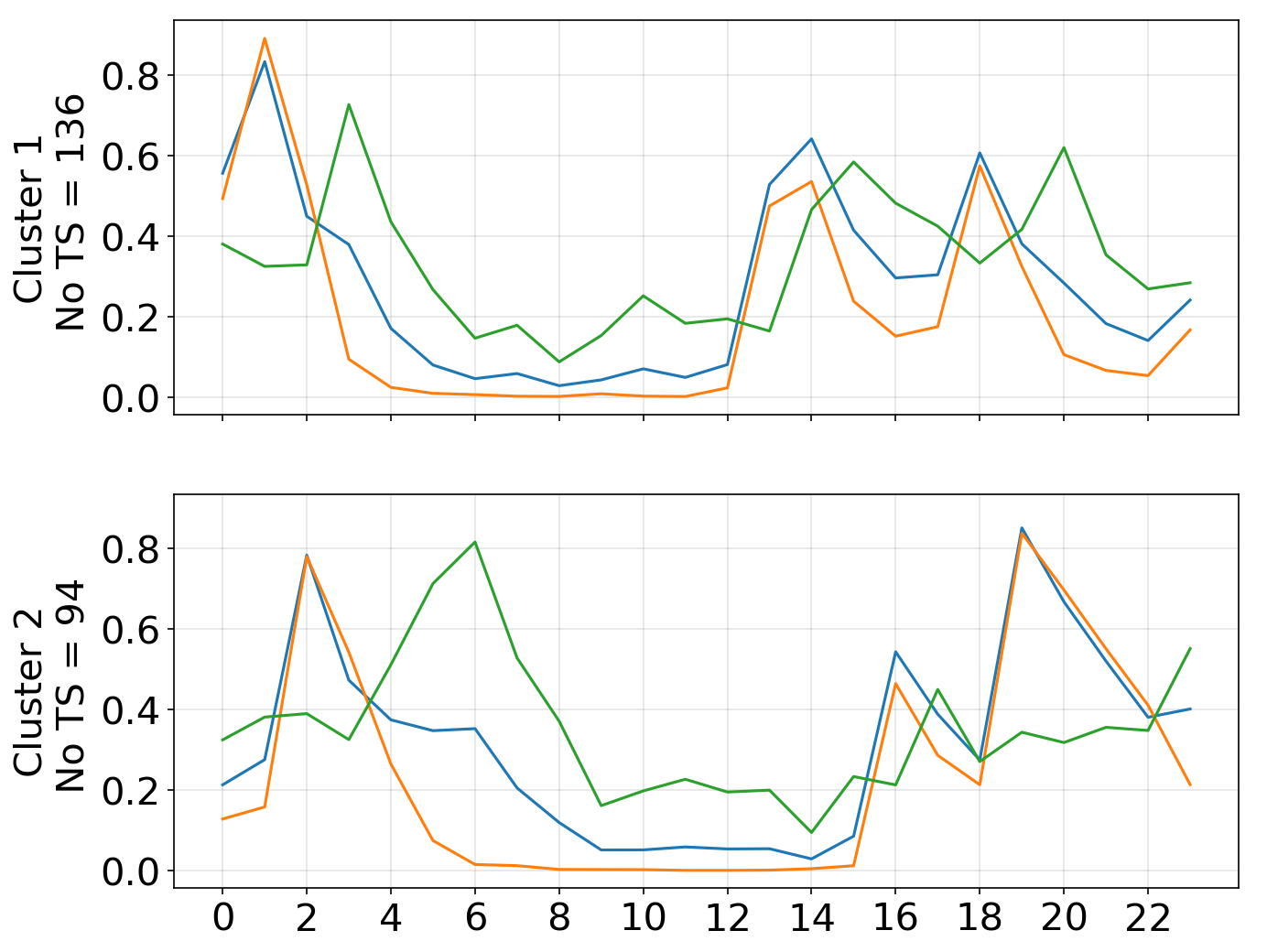}
         \caption{Fold 4}
     \end{subfigure}
     \begin{subfigure}[b]{0.3\textwidth}
         \centering
         \includegraphics[width=\textwidth]{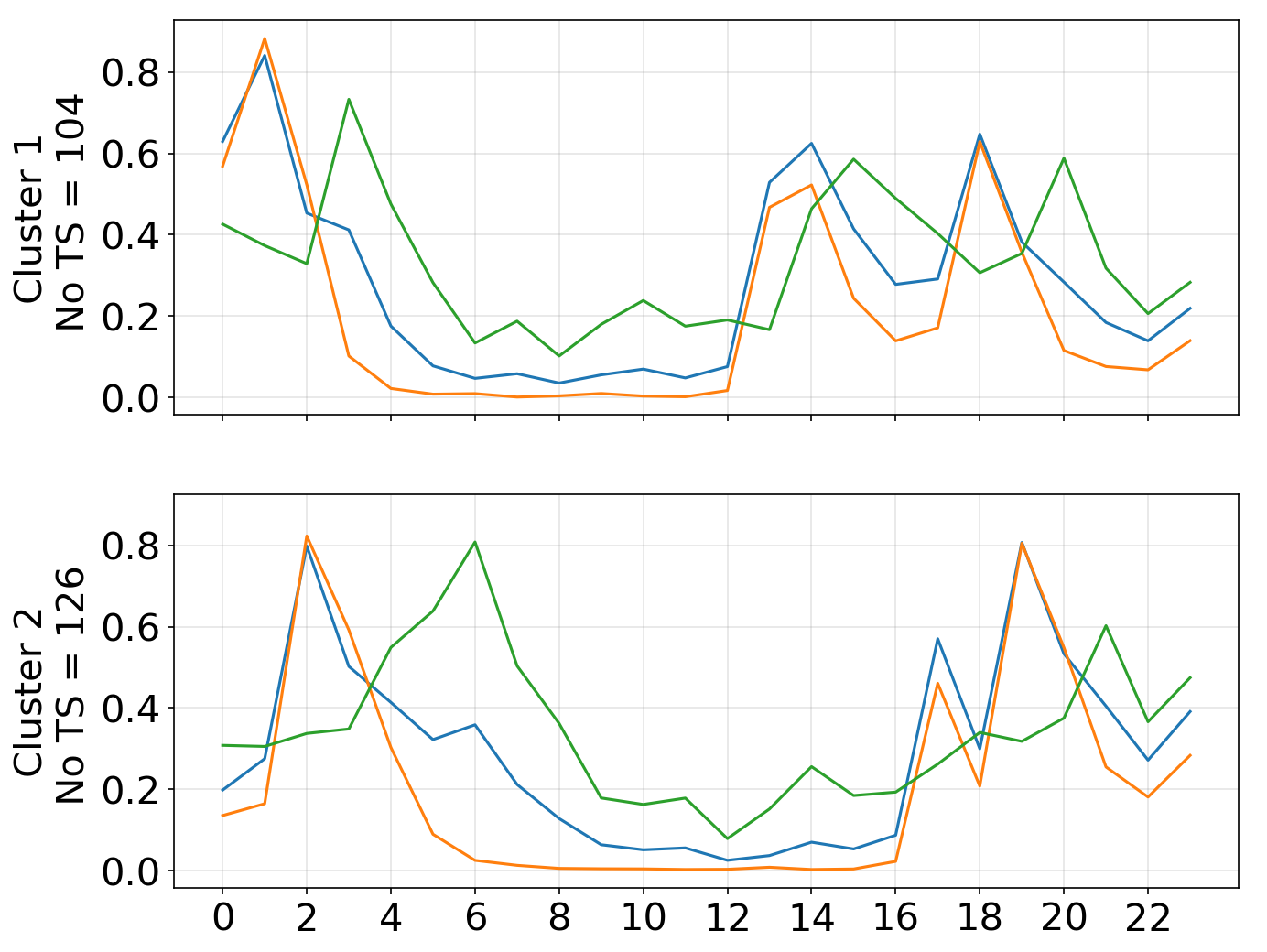}
         \caption{Fold 5}
     \end{subfigure}
     \begin{subfigure}[b]{0.3\textwidth}
         \centering
         \includegraphics[width=\textwidth]{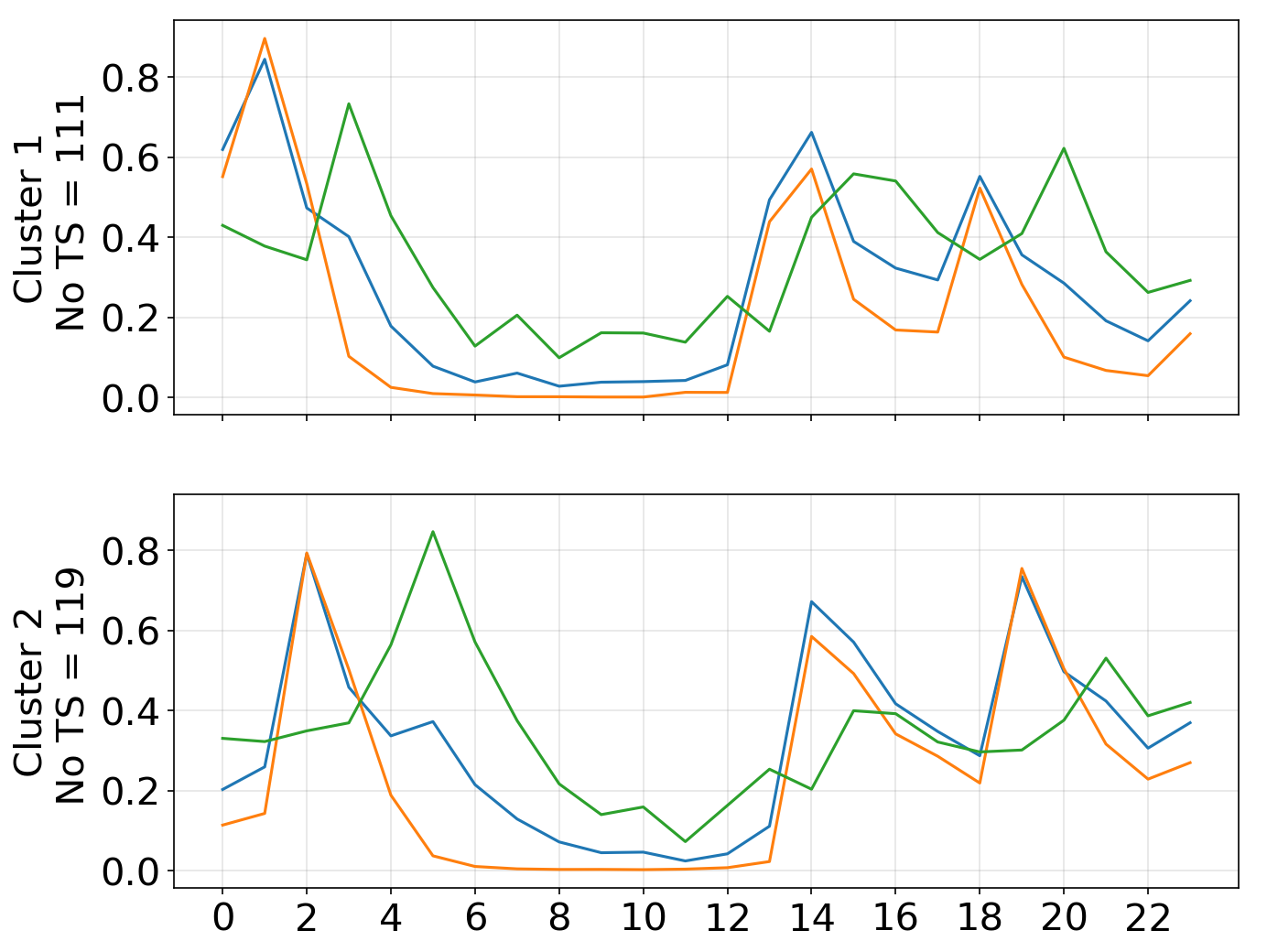}
         \caption{Fold 6}
     \end{subfigure}
     \begin{subfigure}[b]{0.3\textwidth}
         \centering
         \includegraphics[width=\textwidth]{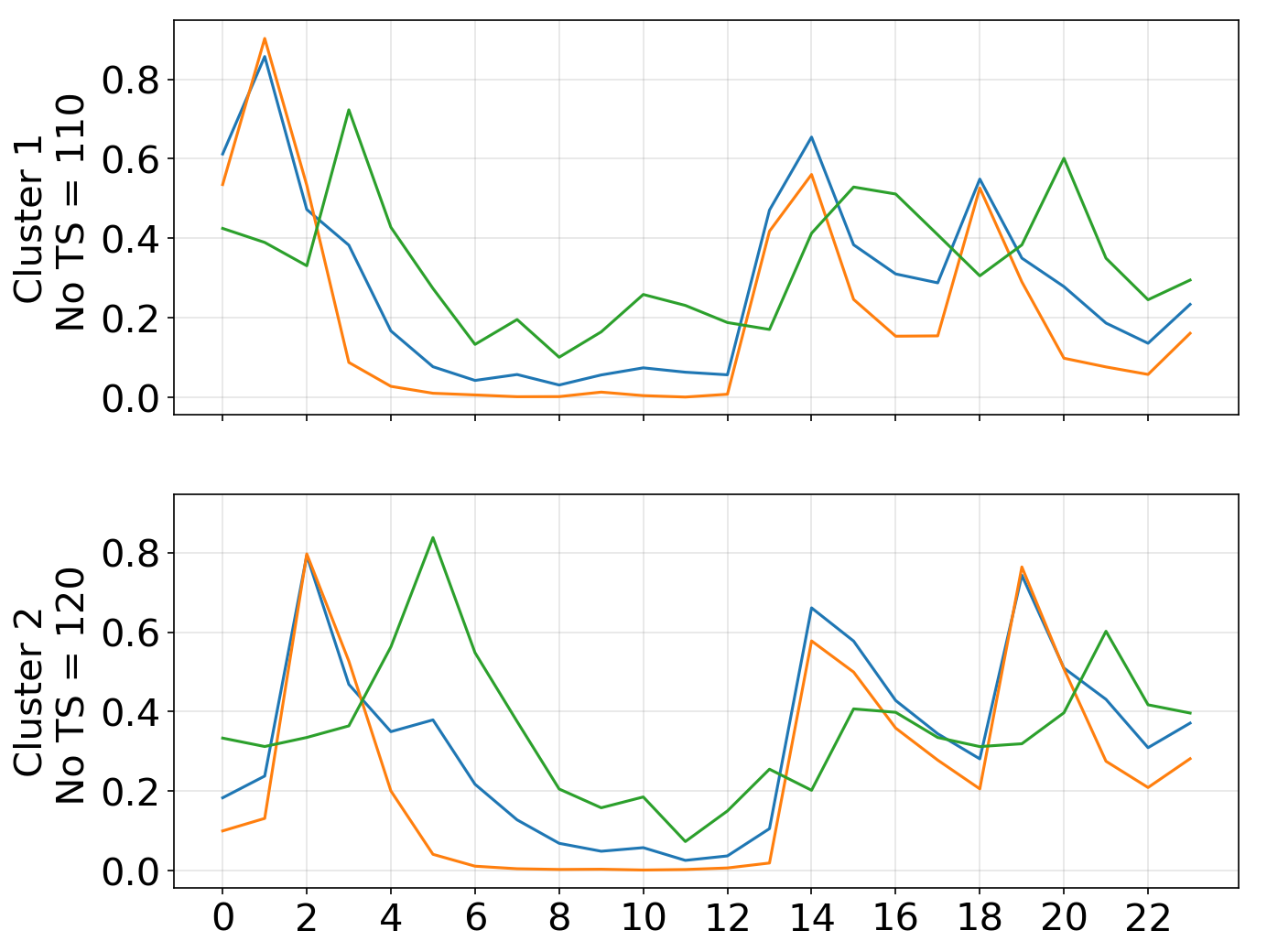}
         \caption{Fold 7}
     \end{subfigure}
     \begin{subfigure}[b]{0.3\textwidth}
         \centering
         \includegraphics[width=\textwidth]{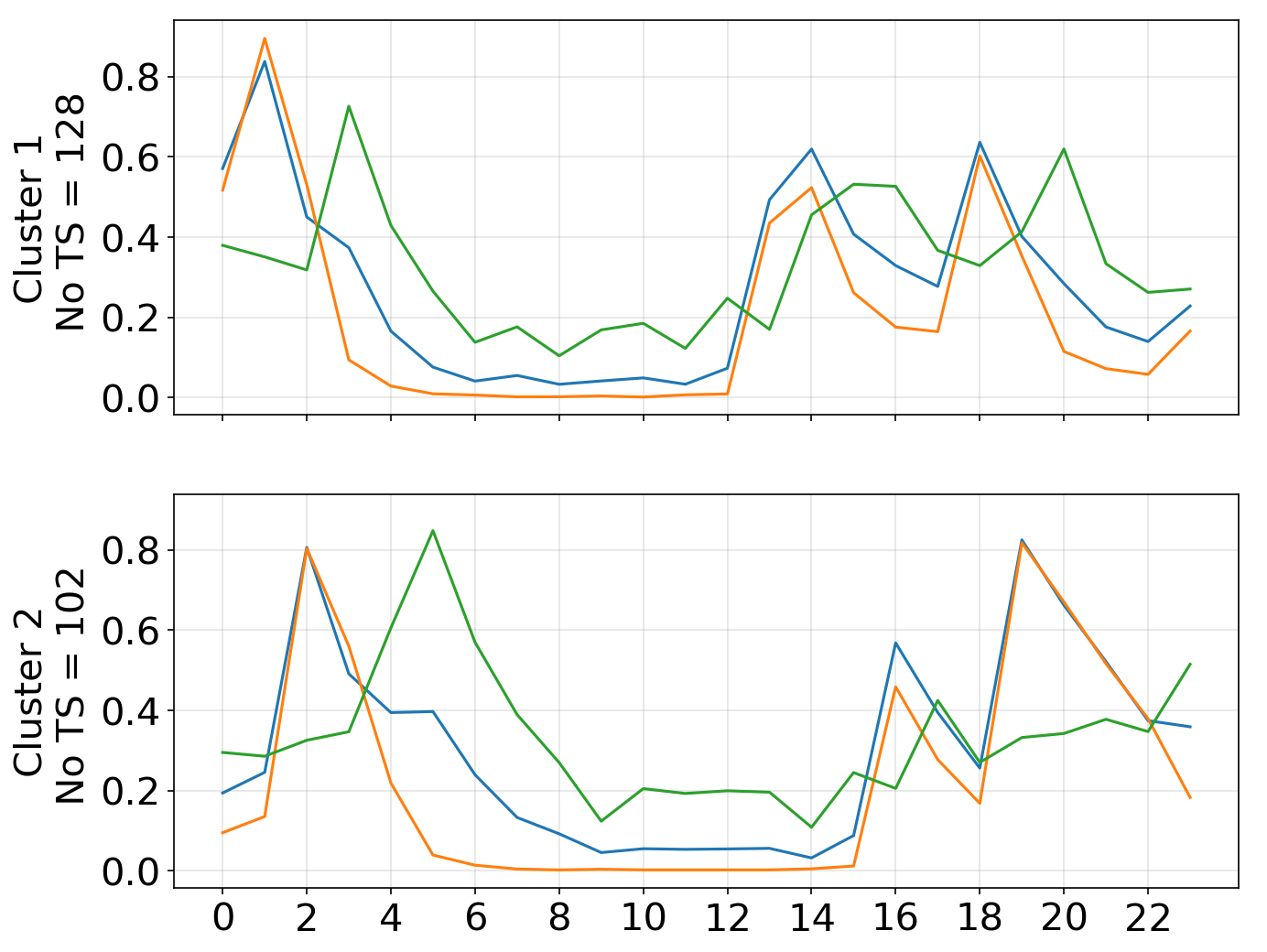}
         \caption{Fold 8}
     \end{subfigure}
     \begin{subfigure}[b]{0.3\textwidth}
         \centering
         \includegraphics[width=\textwidth]{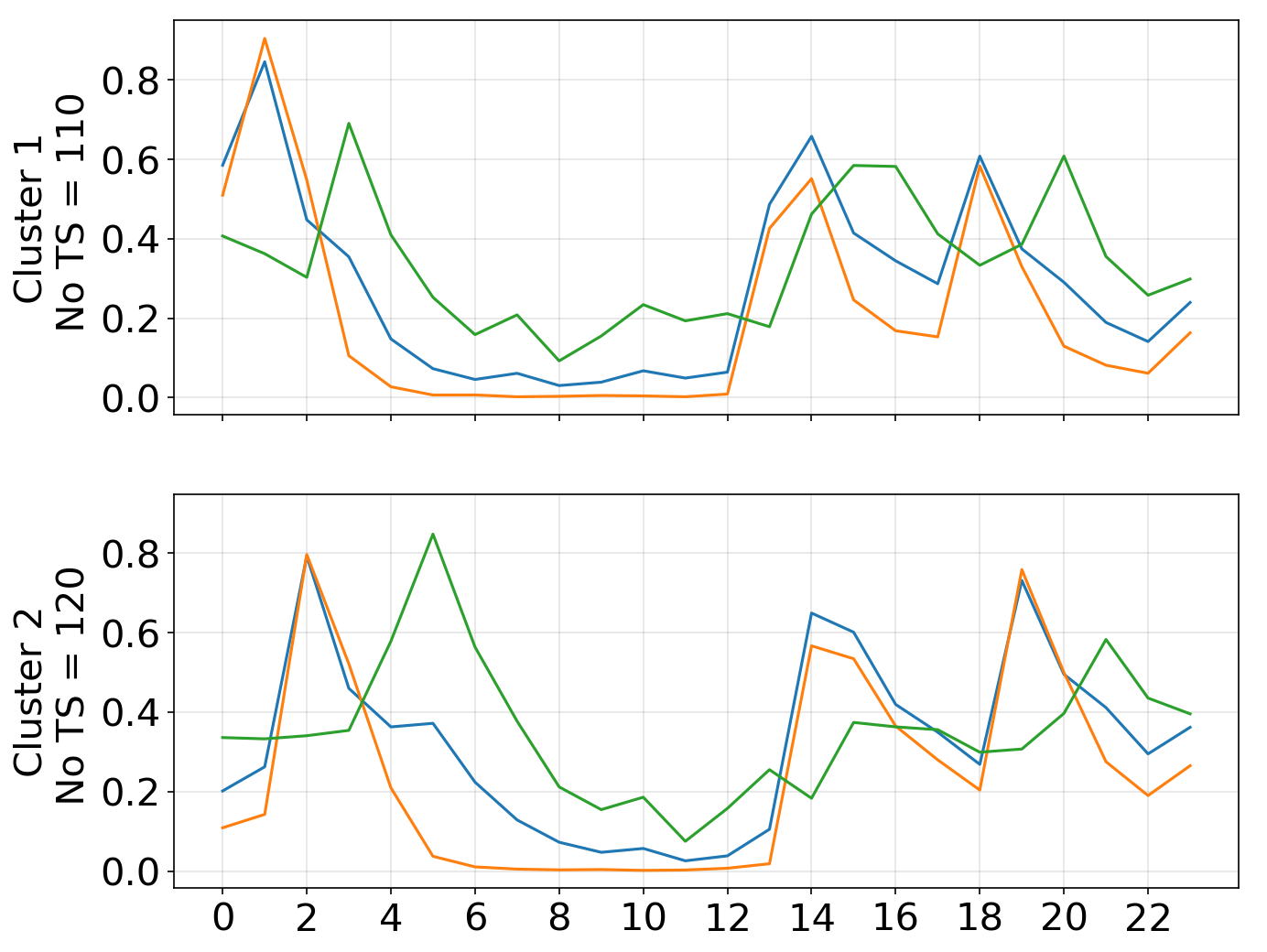}
         \caption{Fold 9}
     \end{subfigure}
     \begin{subfigure}[b]{0.3\textwidth}
         \centering
         \includegraphics[width=\textwidth]{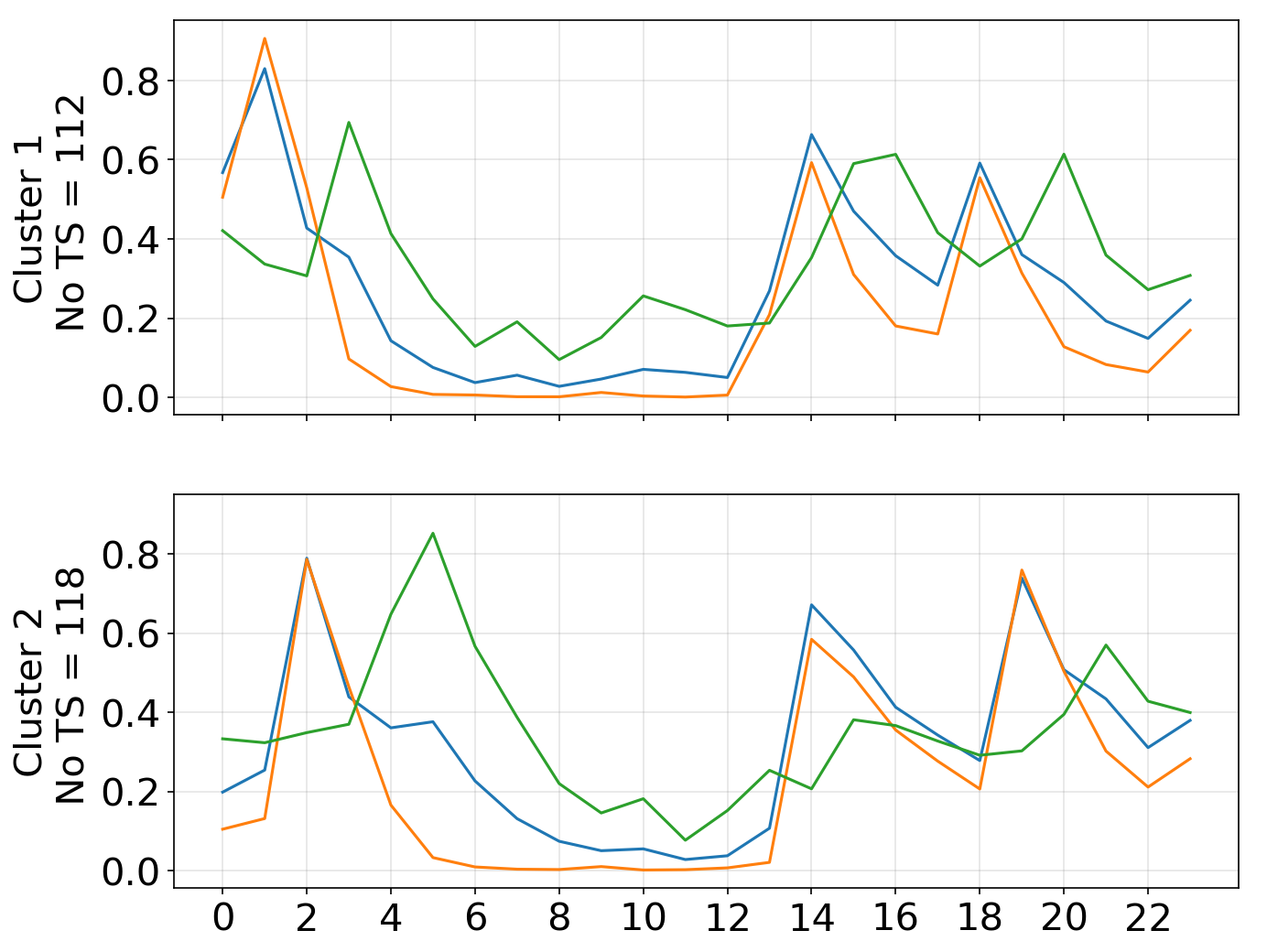}
         \caption{Fold 10}
     \end{subfigure}
     \begin{subfigure}[b]{0.3\textwidth}
         \centering
         \includegraphics[width=\textwidth]{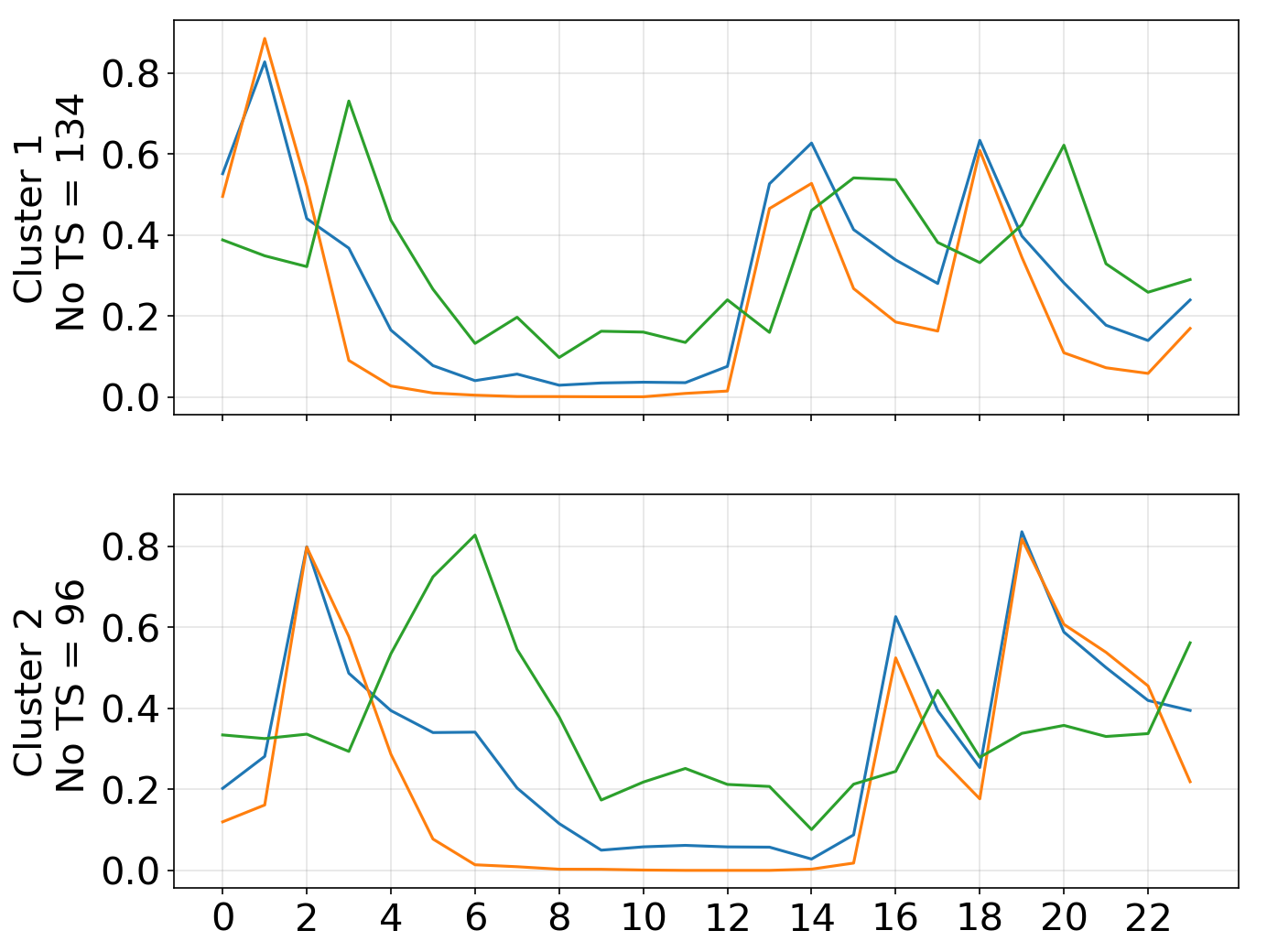}
         \caption{Fold 11}
     \end{subfigure}
     \caption{Barycenters for multivariate k-means clustering of IOB, COB and BG for each of the 11 folds of 230 days out of the 253 days of hourly time series of P1.}
    \label{fig:cross-validated patterns}
\end{figure}

Figure \ref{fig:cross-validated patterns} shows the barycenters for the  original (a) multivariate k-means clustering of the 253 days of hourly data for P1. Sub-figures (b)-(l) show the barycenters when each of the 11 folds (23 time series in each fold) was dropped once and k-means was calculated for the reminding 230 days.

\end{document}